\documentclass{article}

\usepackage[final]{neurips_2024}

\usepackage[utf8]{inputenc} 
\usepackage[T1]{fontenc}    
\usepackage{hyperref}       
\usepackage{url}            
\usepackage{booktabs}       
\usepackage{amsfonts}       
\usepackage{nicefrac}       
\usepackage{microtype}      
\usepackage{microtype}
\usepackage{graphicx}
\usepackage{subcaption}
\usepackage{booktabs} 
\usepackage{amsmath}
\usepackage{placeins}
\usepackage{dblfloatfix}   
\usepackage{wrapfig}
\usepackage[table]{xcolor}
\definecolor{rsgrey}{gray}{0.92}
\usepackage{multirow}
\usepackage{multicol}

\title{Random-Set Graph Neural Networks}

\author{%
    Tommy Woodley \\
    School of Engineering, \\Computing and Mathematics \\
    Oxford Brookes University \\
    \texttt{19204799@brookes.ac.uk} \\
    \And
    Shireen Kudukkil Manchingal\\ 
    School of Engineering, \\Computing and Mathematics \\ Oxford Brookes University\\
    \texttt{smanchingal@brookes.ac.uk}
    \AND
    Matteo Tolloso \\
    Department of Computer Science\\
    University of Pisa \\
    \texttt{matteo.tolloso@phd.unipi.it}
    \And
    Davide Bacciu\\
    Department of Computer Science\\
    University of Pisa\\
    davide.bacciu@unipi.it
    \And
     Fabio Cuzzolin\\
    Oxford Brookes Institute for \\Artificial Intelligence, \\Data Analysis and Systems (AIDAS)\\
    \texttt{fabio.cuzzolin@brookes.ac.uk}   
}

\begin{document}

\maketitle

\begin{abstract}
Uncertainty quantification has become an important factor in understanding the data representations produced by Graph Neural Networks (GNNs). Despite their predictive capabilities being ever useful across industrial workspaces, the inherent uncertainty induced by the nature of the data is a huge mitigating factor to GNN performance. While aleatoric uncertainty is the result of noisy and incomplete stochastic data such as missing edges or over-smoothing, epistemic uncertainty arises from lack of knowledge about a system or model (e.g., a graph's topology or node feature representation), which can be reduced by gathering more data and information. 
In this paper, we propose an original new framework in which node-level epistemic uncertainty is modelled in a belief function (finite random set) formalism. The resulting Random-Set Graph Neural Networks have a belief-function head predicting a random set over the list of classes, from which both a precise probability prediction and a measure of epistemic uncertainty can be obtained. 
Extensive experiments on 9 different graph learning datasets, including real-world autonomous driving benchmarks as such Nuscene and ROAD, demonstrate RS-GNN's superior uncertainty quantification capabilities.
\end{abstract}

\section{Introduction}

Machine learning systems are increasingly deployed in safety-critical domains, where erroneous predictions may lead to severe consequences. In such settings, it is not sufficient for a model to be accurate; it must also be aware of its own limitations; in other words, to know when it does not know \citep{manchingal2022epistemic,manchingal2025epistemic}. A central obstacle to trustworthy deployment is the reliable quantification of \emph{epistemic uncertainty}, the uncertainty arising from limited knowledge about the data-generating process \citep{kendall2017uncertainties, hullermeier2021aleatoric}. 

This challenge is particularly acute for Graph Neural Networks (GNNs) \citep{scarselli2008graph}. Unlike classical learning in which data points are i.i.d. (independent and identically distribute), graph data exhibits strong relational dependencies between instances: uncertainty, therefore, can originate not only from noisy node features but also from structural ambiguity in the graph topology \citep{khan2018uncertain,wang2024uncertainty-graph}. 
A node may be uncertain because its neighborhood is inherently ambiguous (aleatoric uncertainty), or because its structural context is novel and insufficiently represented in the training data (epistemic uncertainty). Topological uncertainties are a factor in signal processing \citep{ceci2020graph}, wireless networks \citep{coon2016topological} and multi-agent systems \citep{yucelen2015control}, to cite just a few examples. Disentangling these sources is essential for robust out-of-distribution (OOD) detection \citep{yang2024generalized} and reliable graph learning \citep{gawlikowski2023survey, valdenegro2022deeper}.

Bayesian approaches provide a principled lens for this decomposition by marginalizing predictions over a posterior distribution of model parameters \citep{malinin2018predictive}. In Graph Neural Networks, Bayesian variants place distributions over weights and, in some cases, over graph structure itself \citep{hasanzadeh2020bayesian, munikoti2023general}. However, Bayesian Graph Neural Networks often incur substantial computational overhead and require carefully specified priors \citep{goan2020bayesian, wang2024uncertainty}. Ensemble-based approaches offer a practical alternative by measuring disagreement across independently trained models \citep{lakshminarayanan2017simple, mallick2022deep}, yet they scale poorly and remain limited to first-order probability representations.
Recent work has explored \emph{credal learning} on graphs \citep{tolloso2025credal}, where models output probability intervals \citep{de1994probability,pearl1988probability,cuzzolin2007properties,cuzzolin2009credal,cuzzolin2022intersection} forming a credal set \citep{caprio2024credal, wang2024credaldeep}. While credal methods provide expressive set-valued predictions and principled uncertainty decompositions, they operate within the space of probability distributions. 
A recent review of uncertainty quantification in neural networks can be found in \citep{wang2025review}.

A recent wave of results in classification, uncertainty quantification and out-of-distribution detection using both \emph{random-set} \citep{manchingal2023random,manchingal2025randomsetneuralnetworksrsnn,manchingal2025epistemic,manchingal2025uncertainty,manchingal2025unifiedevaluationframeworkepistemic} and \emph{credal set} representations \citep{caprio2023imprecise,sale2023volume,caprio2024credal,caprio2025credal,wang2025creinns,wang2024credalwrapper,wang2024credal}, is providing significant evidence that {second-order uncertainty theory} \citep{cuzzolin2021big,cuzzolin2024uncertainty} is a most promising framework for tackling the above challenges (as also supported by the results of the recent Epistemic AI \citep{cuzzolin2024epistemic} Horizon 2020 project\footnote{\url{https://www.epistemic-ai.eu/}}\footnote{\url{https://cordis.europa.eu/project/id/964505/results}}).
Random sets \citep{nguyen,ross86random,smets92TBMrandom,manchingalepistemic,cuzzolin2023reasoning,goutsias97random,ha2009some}, in particular, offer a more fundamental representation of epistemic uncertainty. A random set is a set-valued random variable that assigns probability mass directly to subsets of outcomes \citep{molchanov1997statistical,molchanov1999strong,Molchanov05,shafer1976a}. On finite domains, random sets take the form of \emph{belief functions} \citep{cuzzolin2014belief,cuzzolin14lap,cuzzolin2018belief,cuzzolin2020geometry}, which generalize classical probability measures and naturally induce a convex set of consistent probability distributions (a credal set) \citep{levi80book,cuzzolin2018visions}. 

In the recently proposed Random-Set Neural Networks (RS-NN) \cite{manchingal2025randomsetneuralnetworksrsnn}, predictions are expressed as belief functions over the class list, encoding epistemic uncertainty through the size and geometry of the associated credal set. A point prediction can then be recovered via the pignistic transformation \citep{smets1994transferable}, corresponding geometrically to the center of mass of the credal set.
The approach has recently been extended to large language models \citep{mubashar2025random}, generative adversarial networks \citep{mubashar2026epistemic} and Dirichlet random-set representations in a network's parameter space \citep{sultana2025epistemic}.

While RS-NNs have demonstrated strong performance in image classification and OOD detection, their formulation assumes independent data. Extending random-set learning to graphs, therefore, introduces new conceptual and technical challenges. Message passing in GNNs violates the i.i.d. assumption by construction, and information about a node’s label is progressively gained and lost across layers \citep{fuchsgruber2025uncertainty}. In heterophilic graphs, where connected nodes may belong to different classes, smoothing mechanisms can blur class boundaries and exacerbate epistemic uncertainty \citep{zhu2020beyond, lim2021large}. A random-set formulation must therefore account for the distinctive dynamics of graph information propagation. While message passing in the form of belief functions has been studied in the early Nineties, mostly by Shenoy and Shafer \citep{shenoy1990axioms,shenoy1986propagating,shenoy2023graphical}, the problem has been never tackled from a machine learning perspective.

In this paper, we introduce \emph{Random-Set Graph Neural Networks (RS-GNNs)}, the first framework to extend random-set neural learning to the graph domain. RS-GNN replaces the classical softmax layer of a GNN with a belief-output layer that predicts mass functions over a budgeted collection of focal sets. Each node is thus mapped to a belief function, which is mathematically equivalent to a credal set of probability distributions on the class list. The pignistic probability provides a central prediction, while the geometry of the induced credal set quantifies epistemic uncertainty.

Our approach inherits the theoretical advantages of random sets, namely, the ability to model second-order uncertainty without committing to a precise probability distribution, while adapting them to graph-structured data. By integrating belief-function outputs with message-passing representations, RS-GNN provides a principled and scalable framework for epistemic uncertainty quantification in node classification and OOD detection.

\paragraph{Main Contributions.}
Our contributions are threefold:
(i) we introduce Random-Set Graph Neural Networks (RS-GNNs), extending belief-function learning to graph-structured data;
(ii) we develop a budgeted focal-set selection strategy compatible with graph representations, ensuring scalability to large class spaces;
and (iii) we demonstrate through extensive experiments on a large variety of homophilic and heterophilic benchmarks (including the major real-world autonomous driving datasets \texttt{nuscenes} \citep{caesar2020nuscenes} and \texttt{ROAD} \citep{singh2022road}) that RS-GNN achieves competitive classification performance while providing more reliable epistemic uncertainty estimates and improved OOD detection.

\section{Related Work}
\label{sec:related}
Quantifying uncertainty in Graph Neural Networks (GNNs) \citep{bacciu2020gentle} has emerged as an important and fast-evolving research area, with numerous recent methods aimed at assessing the reliability of GNN predictions \citep{wang2024uncertainty, chen2024uncertainty}. The intrinsic interdependencies among graph nodes pose distinctive challenges, giving rise to a wide range of uncertainty estimation strategies. Existing methods can generally be categorized into three principal families, each characterized by different computational and modeling properties.

The most computationally efficient category is that of \textit{Single Deterministic Models}. The most basic techniques in this group rely on post-hoc heuristics applied to the outputs of a standard GNN, such as adopting the maximum softmax probability as a confidence score or computing predictive entropy. Although straightforward to implement, these methods yield a single, undifferentiated notion of uncertainty and are often susceptible to model miscalibration \citep{guo2017calibration}. A more principled deterministic alternative is evidential deep learning, in which the GNN is trained to predict the parameters of a higher-order Dirichlet distribution. This formulation enables direct modeling of uncertainty over the categorical output space within a single forward pass, offering a more expressive representation of predictive uncertainty \citep{zhao_uncertainty_2020, stadler_graph_2021}. Energy-based approaches have also recently been introduced as a deterministic framework \citep{wu2023energy}. {\citet{wang2021confident} introduced a post-hoc calibration mapping that adjusts the logits of a GNN while preserving their relative ordering across classes. More recently, the Graph Energy-Based Model (GEBM) \citep{fuchsgruber2024energy} derives epistemic uncertainty in a post-hoc manner from the logit-based energy of a pre-trained GNN. This approach regularizes the joint energy to ensure an integrable density and aggregates energy across multiple structural scales, all while retaining strong computational efficiency. In contrast to Bayesian or ensemble-based methods, it does not require retraining and can be readily applied to any logit-based GNN architecture.}

A second prominent paradigm is \textit{Bayesian Graph Neural Networks (Bayesian GNNs)}, which extend the Bayesian framework to graph-structured data. In these models, priors are defined not only over conventional network parameters (e.g., weights) but also over graph-specific elements such as edge connectivity and node feature propagation. For instance, adaptive connection sampling \citep{hasanzadeh2020bayesian} models edges (or adjacency masks) as random variables, while inference techniques such as Monte Carlo Dropout~\citep{gal2016dropout} or Variational Inference \citep{hoffman2013stochastic} are employed to approximate posterior distributions by sampling both model parameters and possible graph realizations at prediction time. This captures uncertainty arising both from the model parameters and from the implications of the graph structure and features. Recent work, such as \citet{munikoti2023general}, highlights how Bayesian approaches on graphs account for uncertainty not only in weights but also in graph topology, node attributes, and edge sampling. Along similar lines, \citet{stadler2021graph} introduced the Graph Posterior Network (GPN), which performs explicit Bayesian posterior updates for predictions over interdependent nodes. While these methods offer a principled framework for modeling uncertainty originating from both model parameters and graph data (structure and features), they typically incur increased computational cost \citep{jia2020efficient} and require careful design of appropriate priors over graph components.

Bridging the divide between theoretical soundness and practical scalability, \textit{Ensemble Methods} have emerged as a widely adopted and effective alternative. This paradigm combines the predictions of multiple independently trained GNNs, with uncertainty quantified through the level of disagreement or variance across individual model outputs \citep{mallick2022deep, busk2023graph}. Although empirically strong—and often competitive with or superior to more sophisticated Bayesian approaches—ensembles impose considerable computational and memory overhead, as they necessitate training and maintaining multiple complete models.

\section{Methodology}
\label{sec:method}

\begin{wrapfigure}{r}{0.52\textwidth}
    \centering
    \includegraphics[width=0.45\textwidth]{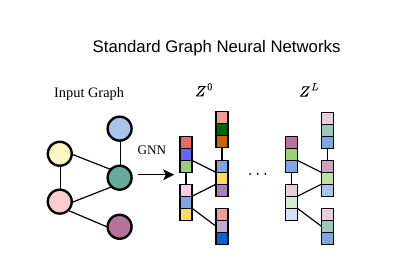}

    \caption{Standard GNN pipeline for node classification, where message passing produces node embeddings followed by a softmax probability vector. }
        \label{fig:GNN}
\end{wrapfigure}

\subsection{Approach: Node-level representation}

\begin{figure}[!ht]
    \centering
        \vspace{-15pt}
    \hspace{-2.3em}
\includegraphics[width=1.05\textwidth]{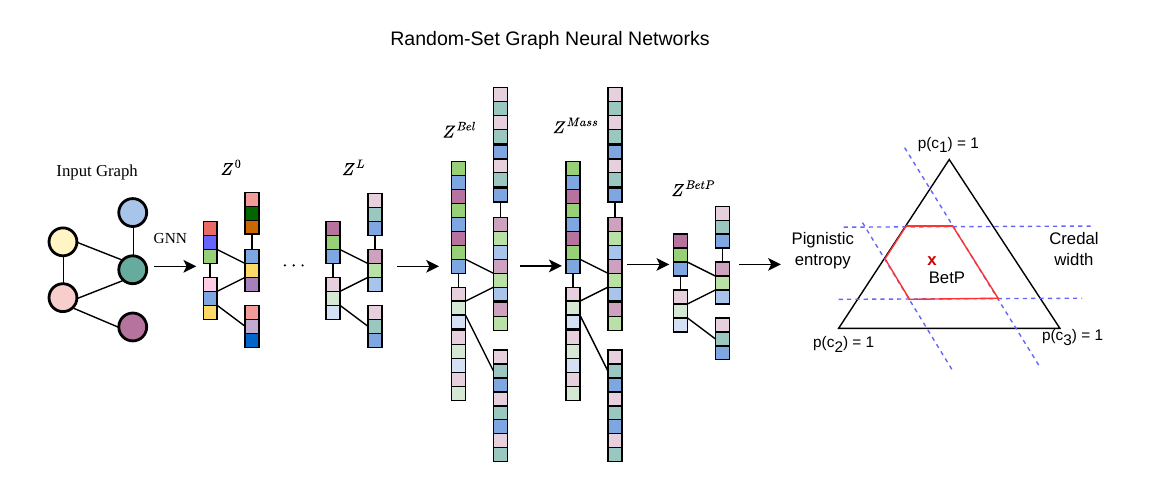}
    \vspace{-15pt}
\caption{
\textbf{Overview of the Random-Set Graph Neural Network (RS-GNN). }
RS-GNN replaces the softmax layer with a belief head that outputs a mass function $m_v$ over a budgeted collection of focal sets $\mathcal{F}$ (singletons plus selected subsets). 
From the predicted mass function we derive (i) a pignistic probability vector $\mathrm{BetP}_v$ for point prediction and (ii) singleton lower and upper probabilities defining the induced credal set. 
On the right is its geometric interpretation in the 3-class probability simplex. The red polygon represents the credal set induced by $m_v$; its center corresponds to the pignistic probability (red cross), whose entropy defines the pignistic entropy. The vertical span of the credal set along the predicted class axis corresponds to the credal set width used as an epistemic uncertainty score.
}
\label{fig:RSGNN}
\end{figure}

We now introduce Random-Set Graph Neural Networks (RS-GNN), a belief-function-based extension of Graph Neural Networks (Fig. \ref{fig:GNN}) for epistemic uncertainty quantification on graph-structured data.
{RS-GNN performs belief-function encoding at the \emph{node level} (Fig. \ref{fig:RSGNN}): each node $v$ is assigned a belief function defined over a shared focal-set budget $\mathcal{F}$. This design follows directly from the node-classification objective, where the relevant uncertainty concerns the class membership of individual nodes given their features and structural context. \emph{Node-level} belief assignments provide localized epistemic uncertainty estimates tied to each node’s neighborhood, enabling fine-grained OOD detection and calibrated decision-making. 
}

{Encoding belief at the whole-graph level would instead capture global model-level uncertainty, which is appropriate for graph-classification tasks but unsuitable for node-level OOD detection. Conversely, encoding beliefs at the edge level would unnecessarily complicate training and interpretation without improving node classification performance. Therefore, assigning belief functions per node provides the most natural and scalable formulation for RS-GNN. In Fig. \ref{fig:RSGNN}, intermediate representations are annotated as $Z^{\text{Bel}}$, $Z^{\text{Mass}}$, and $Z^{\text{BetP}}$ to visually distinguish belief features, mass logits, and pignistic probabilities.}

\subsection{Belief function head}

Let $\mathcal{Y}=\{1,\dots,C\}$ denote the set of node classes and $2^{\mathcal{Y}}$ its power set. In belief-function theory \citep{shafer1976a,cuzzolin14lap}, uncertainty is represented by assigning \emph{belief values} $\mathrm{Bel}(A)$ to subsets $A\subseteq\mathcal{Y}$. Such belief values can be interpreted as lower bounds to the (epistemically uncertain) probability of $A$, while the dual \emph{plausibility values} $\mathrm{Pl}(A) = 1 - \mathrm{Bel}(A^c)$ amont to upper bounds: 
\begin{equation} \label{eq:credal-set}
\mathrm{Bel}(A) \leq P(A) \leq \mathrm{Pl}(A). 
\end{equation}
For a good introduction to belief functions and lower probabilities, please refer to \citep{cuzzolin2024uncertainty}.

In principle, a belief function is defined over the entire power set $2^{\mathcal{Y}}$, containing an exponential number ($2^C$) of subsets. To ensure scalability, RS-GNN operates on a \emph{budgeted focal set} $\mathcal{F}\subseteq 2^{\mathcal{Y}}$ that includes all singleton classes and a selected number of non-singleton subsets. By default, focal sets are generated as all non-empty subsets of the ID class set (full power set). For larger class spaces, the code includes an optional budgeted alternative based on a TSNE+GMM overlap as in RS-NN \citep{manchingal2025randomsetneuralnetworksrsnn}.

In our proposed framework, the RS-GNN backbone computes node representations
\begin{equation}
Z^{l} = \Phi^{l}(Z^{l-1}, \Psi(Z^{l-1})),
\end{equation}
and the final-layer embedding $Z_v^{L}$ is fed into a belief head that directly predicts belief values
\begin{equation}
\hat{\mathrm{Bel}}_v(A_k) \in (0,1), \qquad A_k \in \mathcal{F}.
\end{equation}

Given predicted beliefs, \emph{mass values} $m(A)$ (i.e., probabilities that the correct output is \emph{exactly} the set $A$) can be obtained using the Möbius inverse formula \citep{grabish06moebius,cuzzolin08pricai-moebius,cuzzolin10ida}
\begin{equation}
\hat m_v(A)
=
\sum_{B \subseteq A} (-1)^{|A\setminus B|} \hat{\mathrm{Bel}}_v(B),
\qquad A \in \mathcal{F}.
\end{equation}
To amount to a proper belief function, the mass function so obtained must satisfy non-negativity and normalisation constraints \citep{shafer1976a}, which are enforced through regularisation.
The \emph{pignistic probability} \citep{smets2002decision,smets2005decision} (the center of mass of the credal set (\ref{eq:credal-set}) corresponding to a belief function \citep{cuzzolin2008credal,cuzzolin2010credal}) can then be computed from the derived mass function as follows:
\begin{equation}
\mathrm{BetP}_v(i)
=
\sum_{A_k \ni i} \frac{\hat m_v(A_k)}{|A_k|}.
\end{equation}
The pignistic probability \citep{smets1989constructing,sudano2015pignistic} has the meaning of a central estimate of what the correct categorical probability distribution over the classes is, given the epistemic uncertainty represented by the predicted belief function\footnote{Note that other \emph{probability transforms} are possible, e.g., the orthogonal projection \citep{cuzzolin2007orthogonal}, the intersection probability \citep{cuzzolin2007two} as well as the relative belief \citep{cuzzolin2008dual,cuzzolin2012relative} and plausibility \citep{cobb2006plausibility,cuzzolin2010geometry} of singletons.}.

The predicted class is $\hat y_v = \arg\max_i \mathrm{BetP}_v(i)$.

\subsection{Uncertainty quantification}

In our RS-GNN framework, uncertainty is quantified using two measures. The \textbf{pignistic entropy}
\begin{equation}
H_v = -\sum_{i=1}^{C} \mathrm{BetP}_v(i)\log(\mathrm{BetP}_v(i)+\varepsilon),
\end{equation}
i.e., the Shannon entropy of the predicted pignistic probability,
captures predictive uncertainty. Epistemic uncertainty, instead, is measured via \textbf{credal set width} \citep{sale2023volume}. For singleton sets (individual classes), in particular,
\begin{equation}
\underline{P}_v(i)=\hat m_v(\{i\}),
\qquad
\overline{P}_v(i)=\sum_{A_k : i\in A_k} \hat m_v(A_k),
\end{equation}
and the epistemic uncertainty is the width of the corresponding probability interval (upper bound minus lower bound):
\begin{equation}
W_v=\overline{P}_v(\hat y_v)-\underline{P}_v(\hat y_v).
\end{equation}

\subsection{Loss function and training}

Training is performed directly in belief space. For a node with ground-truth label $y_v$, the target belief encoding is
\begin{equation}
\mathrm{Bel}_v^{\text{target}}(A)=\mathbf{1}\{y_v\in A\}.
\end{equation}
The latter encodes the fact that all sets of classes containing $y_v$ are possible correct predictions, under random-set epistemic uncertainty, so the belief function should assign to them a belief of 1.

The primary loss is binary cross-entropy over focal sets:
\begin{equation}
\mathcal{L}_{\text{Bel}}
=
-\sum_v\sum_{A_k\in\mathcal{F}}
\Big[
\mathrm{Bel}_v^{\text{target}}(A_k)\log(\hat{\mathrm{Bel}}_v(A_k))
+
(1-\mathrm{Bel}_v^{\text{target}}(A_k))\log(1-\hat{\mathrm{Bel}}_v(A_k))
\Big].
\end{equation}

To ensure that the derived mass function is valid, we include regularisation terms that penalise negative mass values and deviations from normalisation:
\begin{equation}
\mathcal{R}(m)
=
\alpha \sum_{A_k\in\mathcal{F}} \max(0,-\hat m_v(A_k))
+
\beta \left|\sum_{A_k\in\mathcal{F}}\hat m_v(A_k)-1\right|.
\end{equation}

The final objective function is therefore:
\begin{equation}
\mathcal{L} = \mathcal{L}_{\text{Bel}} + \mathcal{R}(m).
\end{equation}

\section{Experiments}

\subsection{Datasets}
\label{sec:dataset}

We evaluate our method on a diverse suite of nine benchmark graph datasets, encompassing various domains and structural properties. 

The \texttt{Coauthors} dataset is a computer science co-authorship network where nodes represent authors, connected by an edge if they have co-authored a paper \citep{shchur2018pitfalls}. Node features are derived from paper keywords, and the task is to predict each author's primary field of study. The \texttt{Chameleon} and \texttt{Squirrel} datasets\footnote{While these two datasets have been criticized for having train-test data leakage (derived by duplicate nodes) \citep{platonov2023critical}, they are still widely used in the literature.} are Wikipedia networks where nodes are web pages linked by hyperlinks \citep{rozemberczki2021multi}. Their features are derived from informative nouns on each page, and the classification task is to categorize pages based on their average monthly traffic. \texttt{Reddit2} is a dataset constructed from Reddit posts, where nodes represent individual posts with features generated from text embeddings \citep{hamilton2017inductive}. An edge connects two posts if the same user has commented on both, and the prediction task is to identify the subreddit to which each post belongs. \texttt{ArXiv} is a citation network where nodes are academic papers and directed edges represent citations \citep{hu2020open, ma_revisiting_2024}. Node features are embeddings of the paper's title and abstract, and the objective is to predict the publication year. 
 {The \texttt{Roman Empire} dataset \citep{platonov2023critical} is based on the English Wikipedia article about the Roman Empire, selected for its length and linguistic diversity. Each node corresponds to a (non-unique) word in the text, and two nodes are connected if the corresponding words either appear consecutively in the text or are linked in the dependency tree of a sentence. Node classes correspond to syntactic roles (specifically, the 17 most frequent roles plus an additional class grouping all others). Node features are derived from word embeddings.
The \texttt{Amazon-Ratings} \citep{platonov2023critical} dataset is derived from the Amazon product co-purchasing network. Nodes represent products, and edges connect products that are frequently bought together. The task is to predict the average product rating, grouped into five discrete classes. Node features are computed as the mean of word embeddings of words appearing in the product descriptions.
The \texttt{Cora} dataset \citep{yang2016revisiting} consists of scientific publications represented as nodes, with edges indicating citation links between them. Node features correspond to bag-of-words representations of document content, and the classification task is to assign each publication to its corresponding research topic.}
Finally, \texttt{Patents} is a citation network of U.S. utility patents, where each node is a patent and edges indicate citations between them \citep{leskovec2005graphs, lim2021large, ma_revisiting_2024}. Features are derived from patent metadata, and the task is to predict the patent's grant date.

\subsection{Setting} 
\label{sec:settings}

We address the problem of node-level out-of-distribution (OOD) detection in a transductive setting using a \emph{Leave-Out-Class} strategy during training. 

Let $G = (\mathcal{V}, \mathcal{E}, X)$ denote a single attributed graph, where $\mathcal{V}$ is the set of nodes, $\mathcal{E}$ is the set of edges, and $X \in \mathbb{R}^{|\mathcal{V}| \times d}$ is the node feature matrix. The set of all node classes is partitioned into in-distribution (ID) classes $\mathcal{C}_{ID}$ and out-of-distribution (OOD) classes $\mathcal{C}_{OOD}$, such that $\mathcal{C}_{ID} \cap \mathcal{C}_{OOD} = \emptyset$. 

The node set $\mathcal{V}$ is divided into training, validation, and test subsets. During training, nodes belonging to OOD classes in $\mathcal{C}_{OOD}$ are masked, i.e., their labels are treated as unknown. Consequently, the model is trained to solve a classification problem over only the ID classes $\mathcal{C}_{ID}$. Importantly, in this setting, the true labels of test nodes may include both ID and OOD classes, i.e., the test classes form a superset of the training classes. 

The objective is to learn a model able to classify the in-distribution nodes into their respective classes within $\mathcal{C}_{ID}$ and produce a high uncertainty score for nodes belonging to the OOD classes. 

\subsection{Baselines}
\label{sec:baselines}
We evaluate our proposed credal learning approaches against a suite of established baselines for uncertainty estimation and OOD detection~\citep{ma_revisiting_2024}. These include a family of widely-used post-hoc methods that operate on a single pre-trained GNN, such as the \texttt{Energy}-based score, which is calculated from the pre-softmax logits \citep{liu2020energy}; \texttt{ODIN}, which applies temperature scaling and input perturbations \citep{liang2017enhancing},  {\texttt{CaGCN} \citep{wang2021confident}, a post-hoc calibration methos for logits}, and the \texttt{Mahalanobis} baseline, which measures the distance of a test sample's latent representation from the training data's class-conditional distributions \citep{lee2018simple}. Similarly, we employ a \texttt{K-Nearest Neighbors (KNN)} approach, where uncertainty is derived from the average latent-space distance to the nearest training samples \citep{sun2022out}. We also compare against a method specifically designed for graph data, namely \texttt{GNNSafe} \citep{wu2023energy}  {and the recent \texttt{GEBM} \citep{fuchsgruber2024energy}.}

However, the de-facto reference model against which new uncertainty quantification methods are measured is the \texttt{Classical Ensemble}. Despite its conceptual simplicity, this approach consistently achieves state-of-the-art performance across a wide range of tasks and datasets, making it the target method to outperform \citep{lakshminarayanan2017simple,ovadia2019can, gustafsson2020evaluating, abe2022deep}. Its strength lies in its combination of high performance with practical simplicity: it is easy to implement, scales effectively, and is largely hyperparameter-free, requiring only the independent training of multiple standard models. 

\definecolor{lightgray}{gray}{0.9}

\begin{table*}[h]
\caption{OOD detection performance across all datasets (AUROC $\uparrow$) for baseline methods. Best results are highlighted in \textbf{bold}, second-best are \underline{underlined}, and third-best are shown with a light gray background.}
\label{tab:ood_baselines}
\centering
\resizebox{\textwidth}{!}{%
\begin{tabular}{lcccccc|ccc}
\toprule
Method & Chameleon & Squirrel & ArXiv & Patents & Amazon-Ratings & Roman Empire & Coauthor & Reddit2 & Cora \\
\midrule

Energy & 58.25 & 44.47 & 50.16 & 43.33 & 51.42 & 52.26 & 94.47 & 43.93 & 12.59 \\ 
KNN & 56.81 & 52.79 & 56.86 & 52.55 & \underline{52.46} & 29.32 & 88.23 & 65.63 & 84.67 \\ 
ODIN & 57.98 & 47.28 & 48.16 & 43.21 & 46.23 & 53.98 & 94.17 & 43.09 & 71.65 \\ 
Mahalanobis & 51.82 & 53.79 & 59.52 & 58.72 & 49.66 & 67.26 & 82.49 & 68.98 & 80.26 \\ 
GNNSafe & 50.42 & 35.88 & 35.30 & 27.35 & 49.50 & 50.28 & \colorbox{lightgray}{94.82} & 61.99 & \textbf{88.85} \\ 
Classical ensemble & 74.00/30.22 & 58.32/59.13 & 58.20/65.45 & 48.23/60.35 & 48.51/51.36 & 46.49/55.39 & \underline{95.30}/94.81 & 58.84/\colorbox{lightgray}{72.09} & 75.85/43.84 \\ 
JLDE & 70.06 & 71.06 & 45.35 & 46.74 & 51.72 & 31.01 & 32.37 & 36.55 & 55.36 \\ 
GEBM & 45.41 & 43.00 & 56.26 & 54.49 & 48.79 & 62.62 & 92.20 & 54.49 & \underline{87.15} \\ 
CaGCN & 73.77 & 60.05 & 58.30 & 54.15 & 51.52 & \colorbox{lightgray}{68.03} & \textbf{96.38} & \textbf{91.13} & 81.03 \\ 

\bottomrule
\end{tabular}%
}
\end{table*}

\begin{table*}[h]
\caption{OOD detection performance across all datasets (AUROC $\uparrow$) for credal-based methods and RS-GNN. Best results are highlighted in \textbf{bold}, second-best are \underline{underlined}, and third-best are shown with a light gray background.}
\label{tab:ood_credal}
\centering
\resizebox{\textwidth}{!}{%
\begin{tabular}{lcccccc|ccc}
\toprule
Method & Chameleon & Squirrel & ArXiv & Patents & Amazon-Ratings & Roman Empire & Coauthor & Reddit2 & Cora \\
\midrule

Credal final & \colorbox{lightgray}{76.29}/67.27 & \underline{75.02}/65.85 & 64.65/\colorbox{lightgray}{65.66} & 68.92/\underline{69.73} & 51.89/48.11 & 31.51/\underline{68.48} & 74.80/50.74 & 70.57/69.35 & 43.79/56.17 \\ 
Credal ensemble & 74.98/29.54 & 59.03/53.66 & 58.05/50.97 & 47.41/64.64 & 48.59/51.17 & 45.90/53.06 & 93.85/93.72 & 61.16/57.72 & 75.17/36.42 \\ 
Credal frozen & 48.97/\textbf{85.67} & \colorbox{lightgray}{73.03}/26.96 & 42.36/\textbf{70.22} & 33.29/\colorbox{lightgray}{68.97} & \colorbox{lightgray}{52.41}/48.78 & 68.05/\textbf{69.75} & 93.38/43.78 & 64.69/\underline{73.63} & \colorbox{lightgray}{84.97}/17.33 \\ 
CredalLJ & 72.37/\underline{77.67} & 73.89/\textbf{77.04} & \underline{65.77}/63.79 & \textbf{70.78}/60.06 & \textbf{52.82}/47.27 & 31.45/65.68 & 86.58/70.88 & 66.35/67.31 & 83.26/21.95 \\ 

\midrule
RandomSet GNN & 84.18/\underline{84.10} & 63.57/\textbf{63.56} & \underline{70.21}/69.39 & {49.32}/42.14 & \textbf{51.05}/50.81 & 69.58/69.39 & 85.72/74.56 & 67.38/66.85 & 88.84/88.51 \\ 

\bottomrule
\end{tabular}%
}
\end{table*}

\subsection{Random-Set Graphs for Road Scene Understanding}
\label{sec:rs_road_scene}

We now evaluate Random-Set Graph Neural Networks (RS-GNNs) for road scene understanding in our temporal scene-graph pipeline. In this report, \emph{road scene understanding} is treated as \textbf{risk-aware structured perception}: given a short temporal window, the model must (i) infer semantic categories for traffic entities (node-level recognition) and (ii) quantify uncertainty so that downstream modules (e.g., tracking, behavior prediction, planning) can handle ambiguous or novel situations conservatively rather than acting on overconfident errors.

Formally, each temporal window is represented as a graph
\begin{equation}
\mathcal{G}=(\mathcal{V},\mathcal{E},X),
\end{equation}
where nodes $\mathcal{V}$ represent scene context and traffic entities, edges $\mathcal{E}$ represent spatial/temporal relations, and $X$ are node features. Each labeled agent node $v$ has a semantic class $y_v \in \{0,\dots,C-1\}$, and the primary task is node classification on labeled nodes:
\begin{equation}
\hat y_v = \arg\max_{i \in \{0,\dots,C-1\}} \; p(i \mid v, \mathcal{G}).
\end{equation}
In addition, we evaluate out-of-distribution (OOD) detection under a strict leave-out-class protocol, where a subset of classes is removed from training supervision and only appears at validation/test time. In this setting, good road-scene understanding means not only high ID accuracy, but also \emph{reliable epistemic uncertainty}: OOD nodes should be assigned high uncertainty scores, and ID nodes should not be spuriously flagged.

\subsubsection{Datasets and Temporal Graph Construction}
\label{sec:road_datasets_construction}

We evaluate RS-GNN in two complementary settings: (i) strict leave-out-class OOD within each dataset (ROAD and nuScenes), and (ii) cross-dataset OOD (train on ROAD ID, test OOD on nuScenes). All experiments use the same temporal node-graph pipeline.
Dataset statistics are summarized in Tab. \ref{tab:dataset_stats}.

\textbf{ROAD Dataset.}
Our ROAD benchmark \cite{singh2022road} contains urban and suburban driving scenes with strong variability in illumination, weather, and traffic density. 
Annotations include object-level labels and geometric context sufficient for graph construction at each frame.
The benchmark is particularly suitable for uncertainty analysis because it includes rare classes and rare interaction patterns (e.g., unusual merges, occlusions at intersections, temporary road layouts), which naturally induce epistemic uncertainty.
Following our leave-out-class protocol, a subset of classes is treated as ID during training, while OOD classes appear only at validation/test time.
This setup directly tests whether uncertainty scores assign high epistemic uncertainty to unseen semantic patterns.

\textbf{ROAD temporal node graphs.}
ROAD provides dense temporal annotations with 10 agent categories, 19 action attributes, and 12 location attributes. Each frame contributes one unlabeled scene node and one node per annotated agent. Intra-frame edges are scene-centered (scene node connected bidirectionally to all agents). Temporal edges connect scene-to-scene nodes across adjacent frames and same-\texttt{tube\_uid} agent nodes across adjacent frames; self-loops are added. We use:
\begin{equation}
\text{window size}=4,\qquad \text{window stride}=4,\qquad \text{frame step}=5.
\end{equation}
Node features concatenate box geometry, action one-hot, location one-hot, and node-type indicators; agent-ID features are excluded to prevent leakage. Exported ROAD graphs use 37-dimensional features, with 49{,}123 train graphs, 26{,}302 val graphs, and 26{,}301 test graphs (291{,}197 labeled test nodes).

\textbf{nuScenes.}
We additionally evaluate on \texttt{nuScenes} \citep{caesar2020nuscenes}, a large-scale autonomous-driving benchmark with 1000 scenes and synchronized multi-modal sensing.
The dataset provides:
(i) 6 surround-view cameras,
(ii) 1 LiDAR sensor,
(iii) 5 radars,
(iv) ego-pose and localization metadata,
and (v) map priors for supported regions.
Its geographic and temporal diversity (day/night transitions, weather variability, dense urban traffic) creates a robust testbed for OOD uncertainty behavior in graph models.
We construct scene graphs from fused detections and map elements, using the same graph-building pipeline as for the Road Dataset to ensure a controlled comparison.

\begin{table}[!h]
\centering
\caption{Temporal-graph dataset statistics used in the reported runs.}
\label{tab:dataset_stats}
\begin{tabular}{lrrrr}
\toprule
Dataset & \#Train graphs & \#Val graphs & \#Test graphs & \#Test labeled nodes \\
\midrule
ROAD & 49,123 & 26,302 & 26,301 & 291,197 \\
nuScenes & 307 & 39 & 38 & 6,677 \\
\bottomrule
\end{tabular}
\end{table}
\textbf{nuScenes temporal node graphs.}
For nuScenes, we use 8 object categories:
\begin{equation}
\{\texttt{car},\texttt{truck},\texttt{bus},\texttt{trailer},\texttt{construction\_vehicle},\texttt{pedestrian},\texttt{motorcycle},\texttt{bicycle}\}.
\end{equation}
Temporal windows use
\begin{equation}
\text{window size}=3,\qquad \text{window stride}=1.
\end{equation}
Exported graphs use 10-dimensional node features, with 307 train graphs, 39 val graphs, and 38 test graphs. In the reported balanced run, the evaluated labeled test-node count is 6{,}677.

\subsubsection{Models: What They Do for Road Scene Understanding}
\label{sec:road_models_meaning}

We compare two models that share the same message-passing backbone and differ only in their output representation.

\textbf{Vanilla GNN (softmax head).}
A message-passing backbone computes node embeddings $z_v$, followed by a linear classifier and softmax:
\begin{equation}
p_v(i) = \mathrm{softmax}(W z_v)_i.
\end{equation}
This produces a point probability vector, and uncertainty is typically approximated by scalar confidence heuristics (e.g., maximum softmax probability) or predictive entropy. In closed-set settings this is effective; however, under strict leave-out-class shift, the softmax head still must commit probability mass to singleton classes, which can lead to overconfident errors on unfamiliar (OOD) nodes.

\textbf{RS-GNN (belief head).}
RS-GNN keeps the same backbone but replaces the softmax head with a belief head over a focal-set budget $\mathcal{F}$. For each node $v$, the head outputs beliefs over focal sets (sigmoid activations), which are transformed into masses via M\"obius inversion; a pignistic projection yields a point prediction used for accuracy and calibration metrics. The key difference is representational: RS-GNN can allocate mass to non-singleton focal sets, explicitly encoding \emph{epistemic ambiguity} when the graph evidence is incomplete or conflicting. This is directly beneficial for road-scene understanding, where ambiguity is induced by occlusions, short tracks, rare actors, and novel motion-context patterns: instead of forcing a sharp singleton distribution, RS-GNN can represent partial support for multiple classes while increasing the epistemic uncertainty signal used for OOD detection.

\subsubsection{Training and Leave-Out-Class Protocol}
\label{sec:road_training_protocol}

Both models share the same GATv2 backbone (hidden size 128); differences are only in the output head (softmax vs random sets).

\textbf{Within-dataset leave-out-class OOD.}
For ROAD:
\begin{equation}
\mathcal{Y}_{\mathrm{OOD}}=\{8,9\},\qquad \mathcal{Y}_{\mathrm{ID}}=\{0,\dots,7\}.
\end{equation}
For nuScenes:
\begin{equation}
\mathcal{Y}_{\mathrm{OOD}}=\{6,7\},\qquad \mathcal{Y}_{\mathrm{ID}}=\{0,\dots,5\}.
\end{equation}
Training uses only labeled ID nodes; OOD-labeled nodes are excluded from supervision and used only for uncertainty-based evaluation.

\textbf{Cross-dataset OOD.}
In the ROAD(ID)-nuScenes(OOD) experiment, all ROAD classes are treated as ID for training/evaluation on ROAD, and nuScenes test nodes are treated as OOD at evaluation time.

\subsubsection{Backbone and Training Details}
\label{sec:road_backbone_training_eval}

\textbf{Backbone model.}
All experiments were executed within a dedicated Conda environment, which provided isolated dependency management for Python packages and ensured that all models were evaluated under identical software conditions present in the requirements. Experiments were executed via the command-line supporting all baselines and RS-GNN variants:

\begin{verbatim}
python main.py --dataset <dataset_name> --model <model_name> --count <seeds>
\end{verbatim}

Because training is performed using NeighborLoader-based mini-batching, each forward pass only observes a subgraph of the full dataset. The random set formulation is particularly well-suited to this setting, as it does not require global normalization over the full graph or repeated sampling of model outputs. Each mini-batch produces a self-contained belief assignment over classes of subsets, which remains valid regardless of the sampled neighborhood size. This avoids coupling uncertainty estimation to full-graph computation.

Therefore, this ensures that uncertainty estimation grows with:
\[
\mathcal{O}(2^K)
\]
rather than requiring:
\[
\mathcal{O}(K) \text{ repeated stochastic forward passes (as in ensembles)}.
\]

By constraining uncertainty to a compact set-based representation, the model ensures bounded computational complexity and compatibility with large-scale mini-batch training while still retaining expressive uncertainty estimates suitable for OOD detection.

All experiments in this section use a message-passing Graph Neural Network as the encoder, with \texttt{GATv2} \citep{brody2022how} as the default backbone in our pipeline. GATv2 is a graph attention network variant in which each node aggregates a weighted combination of its neighbors' features, and the weights (attention coefficients) are learned as a function of the interacting node representations. Compared to non-attentive aggregation (e.g., GCN), attention allows the model to down-weight irrelevant or noisy neighbors and emphasize informative relations, which is particularly important in road scene graphs where a node’s neighborhood can contain mixed-quality cues (occlusions, spurious edges, short tracks, and rapidly changing context).

Concretely, our node encoder is a two-layer \texttt{GATv2} message passing network with hidden size 128. The first attention layer uses 4 heads and maps input features to a multi-head hidden representation; the second layer collapses to a single head and returns a 128-dimensional embedding per node. After each layer, we apply a ReLU nonlinearity, and after the first layer we apply dropout with rate 0.2. Denoting node features at layer $l$ by $H^{(l)}$, the encoder can be summarized as
\begin{equation}
H^{(1)} = \sigma\!\left(\mathrm{GATv2}^{(1)}(X,\mathcal{E})\right),
\qquad
H^{(2)} = \sigma\!\left(\mathrm{GATv2}^{(2)}(H^{(1)},\mathcal{E})\right),
\end{equation}
where $\sigma(\cdot)$ is ReLU. The resulting node embedding is $z_v = H^{(2)}_v \in \mathbb{R}^{128}$.

\textbf{Prediction heads (vanilla vs.\ RS-GNN).}
Both models share the same backbone encoder and differ only in the head.

In the \emph{vanilla} model, logits are computed via a linear layer and converted to probabilities with softmax:
\begin{equation}
\ell_v = W z_v,
\qquad
p_v(i) = \mathrm{softmax}(\ell_v)_i.
\end{equation}

In \emph{RS-GNN}, the head outputs focal-set beliefs via a two-layer MLP with sigmoid outputs over a focal-set budget $\mathcal{F}$:
\begin{equation}
\hat{\mathrm{Bel}}_v(A_k) = \mathrm{sigmoid}(g(z_v))_k,
\qquad A_k \in \mathcal{F}.
\end{equation}
Masses over focal sets are recovered by a M\"obius inversion matrix $M$ (precomputed from $\mathcal{F}$):
\begin{equation}
\hat m_v = \hat{\mathrm{Bel}}_v M,
\end{equation}
and pignistic probabilities are computed by multiplying masses with the pignistic matrix $P$ induced by $\mathcal{F}$:
\begin{equation}
\mathrm{BetP}_v = \mathrm{norm}\!\left(\hat m_v P\right),
\end{equation}
where $\mathrm{norm}(\cdot)$ denotes row-wise normalisation to ensure a proper probability vector. Final predictions use $\arg\max_i \mathrm{BetP}_v(i)$, so RS-GNN differs from vanilla primarily in its uncertainty representation rather than in the decision rule.

\textbf{Focal-set selection (how $\mathcal{F}$ is constructed).}
To keep random-set outputs scalable, we do not enumerate all of $2^{\mathcal{Y}}$. Instead, $\mathcal{F}$ contains:
(i) all singleton sets over the full class space, and
(ii) up to 64 additional non-singleton sets (max cardinality 3) selected from warm-up confusion statistics.
This yields 74 focal sets for ROAD (10 singletons + 64 non-singletons) and 72 for nuScenes (8 singletons + 64 non-singletons). The focal-set budgets used for RS-GNN are given in Tab. \ref{tab:focal_sets}.

\begin{table}[!h]
\centering
\caption{Focal-set budgets used for RS-GNN. `K non-singletons' is the number of non-singleton focal sets added to singleton focal sets.}
\label{tab:focal_sets}
\begin{tabular}{lccc}
\toprule
Dataset & \#Classes (total) & K non-singletons & \#Total focal sets (K + singletons) \\
\midrule
ROAD       & 10 (ID=8, OOD=2) & 64 (budgeted pairs/triples) & 74 \\
nuScenes   & 8 (ID=6, OOD=2)  & 64 (budgeted pairs/triples) & 72 \\
\bottomrule
\end{tabular}
\end{table}

\textbf{Training objective.}
Vanilla training minimises cross-entropy on labeled ID nodes. RS-GNN training minimises binary cross-entropy in belief space with mass-validity regularisation. For a labeled node with class $y_v$, the focal-set target is
\begin{equation}
\mathrm{Bel}^{\mathrm{target}}_v(A_k)=\mathbf{1}\{y_v\in A_k\},
\end{equation}
and the training loss is
\begin{equation}
\mathcal{L}
=
\mathrm{BCE}\!\left(\mathrm{Bel}^{\mathrm{target}},\hat{\mathrm{Bel}}\right)
+
\alpha \,\mathbb{E}_v\!\left[ \mathrm{ReLU}\!\left(-\hat m_v\right) \right]
+
\beta \,\mathbb{E}_v\!\left[ \mathrm{ReLU}\!\left(\sum_{A_k\in\mathcal{F}}\hat m_v(A_k)-1\right) \right].
\end{equation}
The first penalty discourages negative mass values; the second discourages total mass exceeding 1 (a relaxed normalisation constraint consistent with our implementation).

\textbf{What counts as a labeled node (label masks).}
In ROAD temporal graphs, each frame includes an unlabeled scene node with label $-1$. We evaluate and train only on nodes with valid labels using a label mask:
\begin{equation}
\mathcal{V}_{\mathrm{lab}}=\{v\in\mathcal{V}: y_v \ge 0\}.
\end{equation}
For strict leave-out-class OOD training, we additionally define the ID mask
\begin{equation}
\mathcal{V}_{\mathrm{ID}}=\{v\in\mathcal{V}_{\mathrm{lab}} : y_v \notin \mathcal{Y}_{\mathrm{OOD}}\},
\end{equation}
and use $\mathcal{V}_{\mathrm{ID}}$ for training updates.

\textbf{Training hyperparameters.}
For ROAD (temporal node graphs), the reported run uses GATv2 with hidden size 128, dropout 0.2, learning rate $10^{-3}$, and regularisation weights $\alpha=\beta=10^{-3}$. We train the vanilla model for 20 epochs and RS-GNN for 30 epochs, with 1 warm-up epoch to build the confusion matrix used for focal-set selection. The batch size is 16.

For nuScenes, the reported balanced run uses the same backbone (GATv2, hidden size 128, dropout 0.2) with learning rate $5\times10^{-4}$, $\alpha=\beta=10^{-3}$, vanilla epochs 60, RS-GNN epochs 80, and warm-up epochs 3. Because nuScenes exhibits strong class imbalance in our strict leave-out split, we additionally use class-weighted losses and label smoothing with smoothing factor 0.02. The batch size is 16. Core experimental hyperparameters for the hold-out-class runs are listed in Tab. \ref{tab:hyperparams}.

\begin{table}[!h]
\centering
\caption{Core hyperparameters used in the two hold-out-class runs.}
\label{tab:hyperparams}
\begin{tabular}{lcc}
\toprule
Hyperparameter & ROAD & nuScenes \\
\midrule
Backbone & GATv2 (2 layers) & GATv2 (2 layers) \\
Hidden dim & 128 & 128 \\
Attention heads & 4 (first), 1 (second) & 4 (first), 1 (second) \\
Dropout & 0.2 & 0.2 \\
Batch size & 16 & 16 \\
Learning Rate & $1\times10^{-3}$ & $5\times10^{-4}$ \\
Epochs & 30 & 80 \\
$\alpha$, $\beta$ & $10^{-3},10^{-3}$ & $10^{-3},10^{-3}$ \\
\bottomrule
\end{tabular}
\end{table}

\subsubsection{Evaluation Protocol}
\label{sec:road_backbone_training_eval_protocol}

We evaluate both models on labeled test nodes. For vanilla, class probabilities are softmax outputs $p_v$. For RS-GNN, class probabilities are pignistic probabilities $\mathrm{BetP}_v$.

\textbf{Entropy-based scores.}
\begin{equation}
H_v = -\sum_{i=0}^{C-1} p_v(i)\log\!\big(p_v(i)+\varepsilon\big), \qquad
s_v^{\mathrm{MSP}} = 1-\max_i p_v(i).
\end{equation}
For binary OOD detection ($t_v=\mathbf{1}\{y_v\in\mathcal{Y}_{\mathrm{OOD}}\}$), we report AUROC/AUPRC/FPR95 using $H_v$ and $s_v^{\mathrm{MSP}}$.

\textbf{Credal-width score (RS-GNN).}
Let $\hat y_v=\arg\max_i \mathrm{BetP}_v(i)$, and let $\hat m_v(A)$ be the predicted mass over the selected focal budget $\mathcal{F}$.
For singleton class $i$:
\begin{equation}
\underline P_v(i)=\hat m_v(\{i\}),\qquad
\overline P_v(i)=\sum_{A:\,i\in A}\hat m_v(A).
\end{equation}
We define predicted-class credal width:
\begin{equation}
W_v=\overline P_v(\hat y_v)-\underline P_v(\hat y_v).
\end{equation}
We report AUROC/AUPRC/FPR95 for $W_v$, plus mean $W_v$ on ID and OOD subsets.

\textbf{Calibration and proper scoring rules.}
We report ECE (15 bins), multiclass NLL, and multiclass Brier score:
\begin{equation}
\mathrm{ECE}=\sum_{b=1}^{B}\frac{|S_b|}{N}\left|\mathrm{acc}(S_b)-\mathrm{conf}(S_b)\right|,\quad B=15,
\end{equation}
\begin{equation}
\mathrm{NLL}=-\frac{1}{N}\sum_{v=1}^{N}\log\!\big(p_v(y_v)+\varepsilon\big),
\end{equation}
\begin{equation}
\mathrm{Brier}=\frac{1}{N}\sum_{v=1}^{N}\sum_{i=0}^{C-1}\left(p_v(i)-\mathbf{1}\{y_v=i\}\right)^2.
\end{equation}

\subsubsection{Results}
\label{sec:road_results_full}

We evaluate three aspects of performance:
(i) in-distribution (ID) accuracy,
(ii) calibration (ECE, NLL, Brier),
and (iii) out-of-distribution (OOD) detection under strict leave-out-class and cross-dataset shift.

\textbf{In-distribution performance.}
Tab. \ref{tab:id_core_metrics} reports ID metrics.  
On \textit{ROAD held-out OOD}, both models achieve comparable accuracy (0.754 vs.\ 0.746), but RS-GNN improves calibration substantially (ECE 0.132 vs.\ 0.178; lower NLL and Brier), indicating better probabilistic reliability even without sacrificing accuracy. On \textit{nuScenes}, the difference is more pronounced: RS-GNN improves accuracy from 0.418 to 0.593 and significantly reduces NLL (7.047 → 2.079). This suggests that belief-based modeling stabilizes training under stronger class imbalance and shorter temporal windows.
In the cross-dataset \textit{ROAD(ID)-nuScenes(OOD)} setting, RS-GNN slightly improves ID accuracy (0.770 vs.\ 0.751), though vanilla softmax achieves slightly better ECE and NLL on ROAD alone. This reflects the expected trade-off: random-set heads prioritize uncertainty separation over pure closed-set likelihood optimisation.

\begin{table*}[!h]
\centering
\caption{In-distribution (ID) performance across ROAD and nuScenes settings.
Metrics are reported on ID nodes only. For ROAD(ID)-NuScenes(OOD), ID metrics are computed on ROAD test nodes.
}
\label{tab:id_core_metrics}
\small
\vspace{-8pt}
\resizebox{\textwidth}{!}{
\begin{tabular}{l l l l l c c c c}
\toprule
Setting & Dataset & Split & Model & Uncertainty & Acc ($\uparrow$) & ECE ($\downarrow$) & NLL ($\downarrow$) & Brier ($\downarrow$) \\
\midrule
\multirow{2}{*}{ROAD held-out OOD}
& \multirow{2}{*}{ROAD}
& ID: $\{0,\dots,7\}$
& Vanilla & Softmax entropy & \textbf{0.754} & 0.178 & 1.849 & 0.431 \\
& & OOD: $\{8,9\}$ & \cellcolor{rsgrey}RS-GNN & \cellcolor{rsgrey}Pignistic entropy & \cellcolor{rsgrey}0.746 & \cellcolor{rsgrey}\textbf{0.132} & \cellcolor{rsgrey}\textbf{1.478} & \cellcolor{rsgrey}\textbf{0.384} \\
\midrule
\multirow{2}{*}{NuScenes held-out OOD}
& \multirow{2}{*}{NuScenes}
& ID: $\{0,\dots,5\}$
& Vanilla & Softmax entropy & {0.418} & 0.452 & 7.047 & 0.905 \\
& & OOD: $\{6,7\}$ & \cellcolor{rsgrey}RS-GNN & \cellcolor{rsgrey}Pignistic entropy & \cellcolor{rsgrey}\textbf{0.593} & \cellcolor{rsgrey}\textbf{0.422} & \cellcolor{rsgrey}\textbf{2.079} & \cellcolor{rsgrey}\textbf{0.875} \\
\midrule
\multirow{2}{*}{ROAD(ID) vs NuScenes(OOD)}
& \multirow{2}{*}{ROAD}
& ID: all ROAD classes
& Vanilla & Softmax entropy & 0.751 & \textbf{0.122} & \textbf{1.035} & \textbf{0.381} \\
& & OOD: NuScenes  & \cellcolor{rsgrey}RS-GNN & \cellcolor{rsgrey}Pignistic entropy & \cellcolor{rsgrey}\textbf{0.770} & \cellcolor{rsgrey}0.157 & \cellcolor{rsgrey}1.837 & \cellcolor{rsgrey}0.398 \\
\bottomrule
\end{tabular}
}
\end{table*}

\textbf{OOD uncertainty separation.}
Tab. \ref{tab:ood_uncertainty_auroc} shows entropy and credal-width behavior.
\textbf{(i) ROAD held-out OOD.}
Vanilla entropy separation is weak (AUROC 0.425).  
RS-GNN substantially improves entropy-based OOD detection (AUROC 0.662) and additionally provides a credal-width signal (AUROC 0.342), demonstrating that belief mass spreads meaningfully increase under unseen classes.
\textbf{(ii) nuScenes held-out OOD.}
Vanilla softmax entropy collapses near zero for both ID and OOD nodes, yielding near-random detection (AUROC 0.463).  
RS-GNN maintains non-trivial entropy levels and improves detection (entropy AUROC 0.500; credal-width AUROC 0.521). Although absolute performance remains moderate, uncertainty does not degenerate as in the softmax model.
\textbf{(iii) Cross-dataset shift (ROAD-nuScenes).}
This is the most severe scenario. Vanilla entropy completely collapses (AUROC $\approx$ 0.001), meaning the softmax model assigns high confidence to foreign-domain nodes.  
RS-GNN improves detection (entropy AUROC 0.053; credal-width AUROC 0.083), showing partial robustness, though cross-dataset generalisation remains fundamentally challenging.

\begin{table*}[!h]
\centering
\caption{
Out-of-distribution uncertainty and detection evaluation across ROAD and nuScenes settings.
For Vanilla, uncertainty is softmax entropy; for RS-GNN, uncertainty is pignistic entropy with credal-width metrics where available.
}
\label{tab:ood_uncertainty_auroc}
\small
\vspace{-8pt}
\resizebox{\textwidth}{!}{
\begin{tabular}{l l c c c c c c c c}
\toprule
\multirow{2}{*}{Setting} &
\multirow{2}{*}{Model} &
\multicolumn{2}{c}{In-distribution (iD)} &
\multicolumn{2}{c}{Out-of-distribution (OOD)} &
\multicolumn{4}{c}{OOD Detection Metrics} \\
\cmidrule(lr){3-4}\cmidrule(lr){5-6}\cmidrule(lr){7-10}
& &
\begin{tabular}[c]{@{}c@{}}Entropy\\(ID) ($\downarrow$)\end{tabular} &
\begin{tabular}[c]{@{}c@{}}Credal\\(ID) ($\downarrow$)\end{tabular} &
\begin{tabular}[c]{@{}c@{}}Entropy\\(OOD) ($\uparrow$)\end{tabular} &
\begin{tabular}[c]{@{}c@{}}Credal\\(OOD) ($\uparrow$)\end{tabular} &
\begin{tabular}[c]{@{}c@{}}AUROC\\(Entropy) ($\uparrow$)\end{tabular} &
\begin{tabular}[c]{@{}c@{}}AUPRC\\(Entropy) ($\uparrow$)\end{tabular} &
\begin{tabular}[c]{@{}c@{}}AUROC\\(Credal) ($\uparrow$)\end{tabular} &
\begin{tabular}[c]{@{}c@{}}AUPRC\\(Credal) ($\uparrow$)\end{tabular} \\
\midrule

\multirow{2}{*}{\begin{tabular}[c]{@{}l@{}}ROAD held-out OOD\\($\mathcal{Y}_{OOD}=\{8,9\}$)\end{tabular}}
& Vanilla & {0.180} & -- & {0.226} & -- & {0.425} & {0.208} & -- & -- \\
& \cellcolor{rsgrey}RS-GNN & \cellcolor{rsgrey}\textbf{0.158} & \cellcolor{rsgrey}\textbf{0.094} & \cellcolor{rsgrey}\textbf{0.354} & \cellcolor{rsgrey}\textbf{0.017} & \cellcolor{rsgrey}\textbf{0.662} & \cellcolor{rsgrey}\textbf{0.327} & \cellcolor{rsgrey}\textbf{0.342} & \cellcolor{rsgrey}\textbf{0.186} \\
\midrule

\multirow{2}{*}{\begin{tabular}[c]{@{}l@{}}nuScenes held-out OOD\\($\mathcal{Y}_{OOD}=\{6,7\}$)\end{tabular}}
& Vanilla & \textbf{0.000124} & -- & 0.000098 & -- & 0.463 & 0.071 & -- & -- \\
& \cellcolor{rsgrey}RS-GNN & \cellcolor{rsgrey}0.045 & \cellcolor{rsgrey}\textbf{2.057} & \cellcolor{rsgrey}\textbf{2.079} & \cellcolor{rsgrey}\textbf{2.061} & \cellcolor{rsgrey}\textbf{0.500} & \cellcolor{rsgrey}\textbf{0.077} & \cellcolor{rsgrey}\textbf{0.521} & \cellcolor{rsgrey}\textbf{0.076} \\
\midrule

\multirow{2}{*}{\begin{tabular}[c]{@{}l@{}}ROAD (ID)\\vs NuScenes (OOD)\end{tabular}}
& Vanilla & 0.329 & -- & $2.48\times10^{-10}$ & -- & 0.001 & 0.470 & -- & -- \\
& \cellcolor{rsgrey}RS-GNN & \cellcolor{rsgrey}\textbf{0.200} & \cellcolor{rsgrey}\textbf{0.065} & \cellcolor{rsgrey}\textbf{1.86}$\times$\textbf{$10^{-7}$} & \cellcolor{rsgrey}\textbf{0.121} & \cellcolor{rsgrey}\textbf{0.053} & \cellcolor{rsgrey}\textbf{0.500} & \cellcolor{rsgrey}\textbf{0.083} & \cellcolor{rsgrey}\textbf{0.500} \\
\bottomrule
\end{tabular}
}
\vspace{-4mm}
\end{table*}

Across settings, three consistent patterns emerge:
\begin{itemize}
\item Softmax heads maintain competitive ID accuracy but struggle to express epistemic uncertainty when encountering unseen classes.
\item RS-GNN improves calibration and preserves uncertainty signals under class removal and moderate domain shift.
\item Under extreme cross-dataset shift, both models degrade, but RS-GNN avoids the complete overconfidence collapse observed in softmax.
\end{itemize}

For road-scene understanding, this is crucial:  
softmax models tend to produce confident singleton predictions even when graph evidence is insufficient, whereas RS-GNN can distribute mass across focal sets, increasing entropy and credal width in ambiguous or novel scenarios.  

In safety-critical pipelines, this translates to better downstream gating of tracking and planning modules, reducing the risk of overconfident errors on rare actors or unfamiliar motion-context patterns.

\begin{figure}[!ht]
    \centering
        \vspace{-15pt}
    \hspace{-2.3em}
\includegraphics[width=1.05\textwidth]{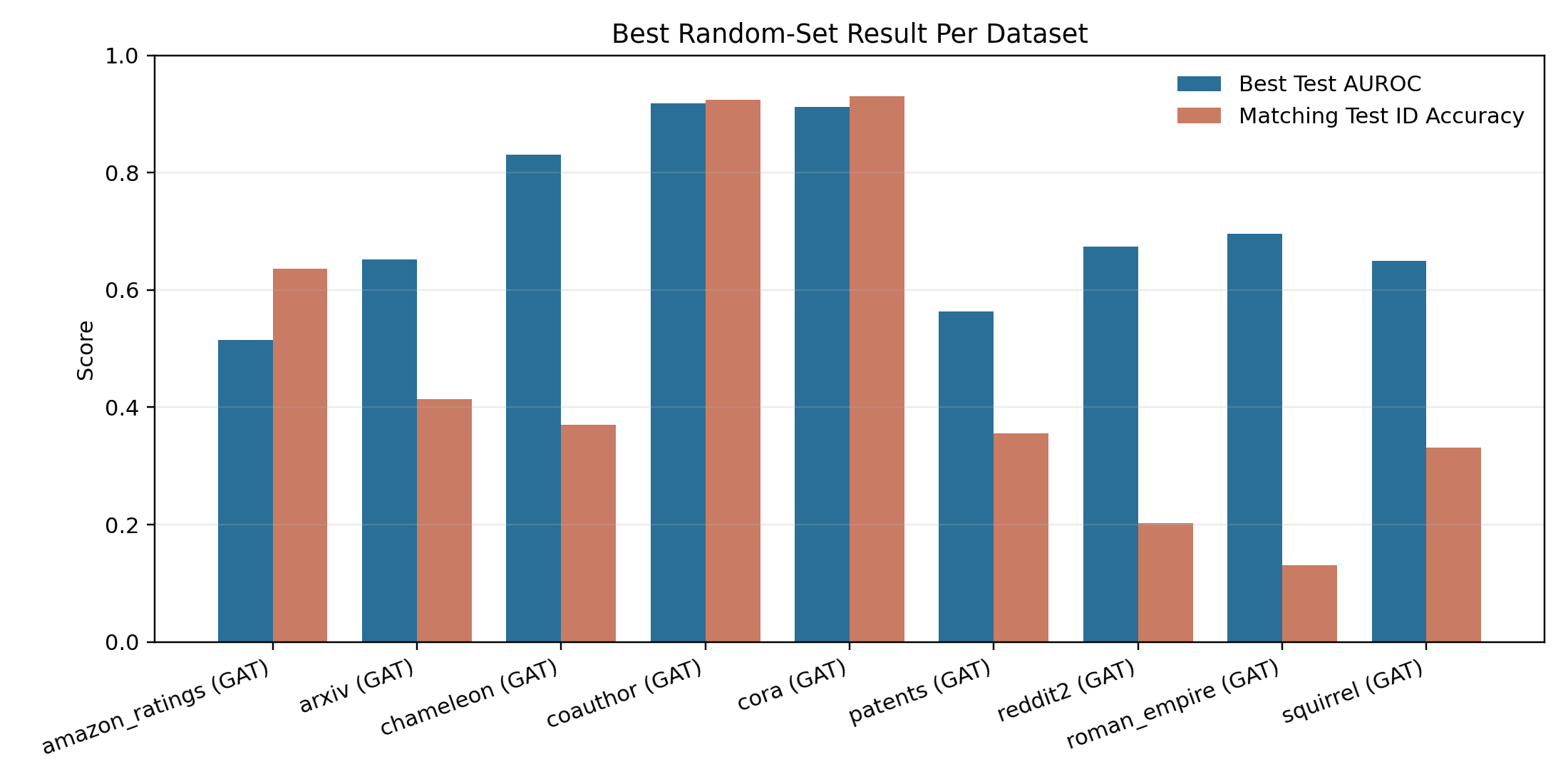}
    \vspace{-15pt}
    \caption{This figure presents the best-performing RS-GNN configuration for each dataset, providing a clear cross-dataset comparison of model performance.}
    \label{fig:best_per_dataset}
    \end{figure}

These results in \ref{fig:best_per_dataset} further show that the random-set model should not be judged solely by its single best run. While the best-performing configuration for each dataset is informative, the distribution of results across runs provides a more reliable picture of model stability. For example, on \texttt{Patents} and \texttt{reddit2}, the spread in AUROC values across experiments indicates that uncertainty quality can be highly sensitive to the chosen configuration. This suggests that the method has the capacity to produce strong OOD separation, but that such performance is not equally robust under all hyperparameter settings. In contrast, where a dataset exhibits a tighter clustering of results, the method can be interpreted as behaving more consistently, even if the absolute best score is not the highest among all benchmarks. This distinction is important: a method that performs well only under a narrow range of settings should be characterised differently from one whose performance is more stable across repeated trials.

An important feature of the present experiments is that focal-set construction differs across datasets. For datasets with a small number of ID classes, such as \texttt{patents}, \texttt{cora}, \texttt{squirrel}, \texttt{chameleon}, and \texttt{roman empire}, the model uses the full power set of non-empty focal sets. In these cases, the random-set representation remains relatively direct, and uncertainty can be expressed across all class subsets. By contrast, for datasets with a larger number of classes, such as \texttt{reddit2} and others with more complex label spaces, the full power set becomes computationally impractical, and a budgeted focal-set construction is used instead. This difference is not merely an implementation detail; it has methodological implications. The results suggest that the quality of uncertainty estimates depends not only on the learned graph representation, but also on the expressiveness of the focal-set family through which mass is distributed. In other words, the uncertainty mechanism itself forms part of the model capacity.

\begin{figure}[!ht]
    \centering
    \hspace{-2.3em}
\includegraphics[width=1.05\textwidth]{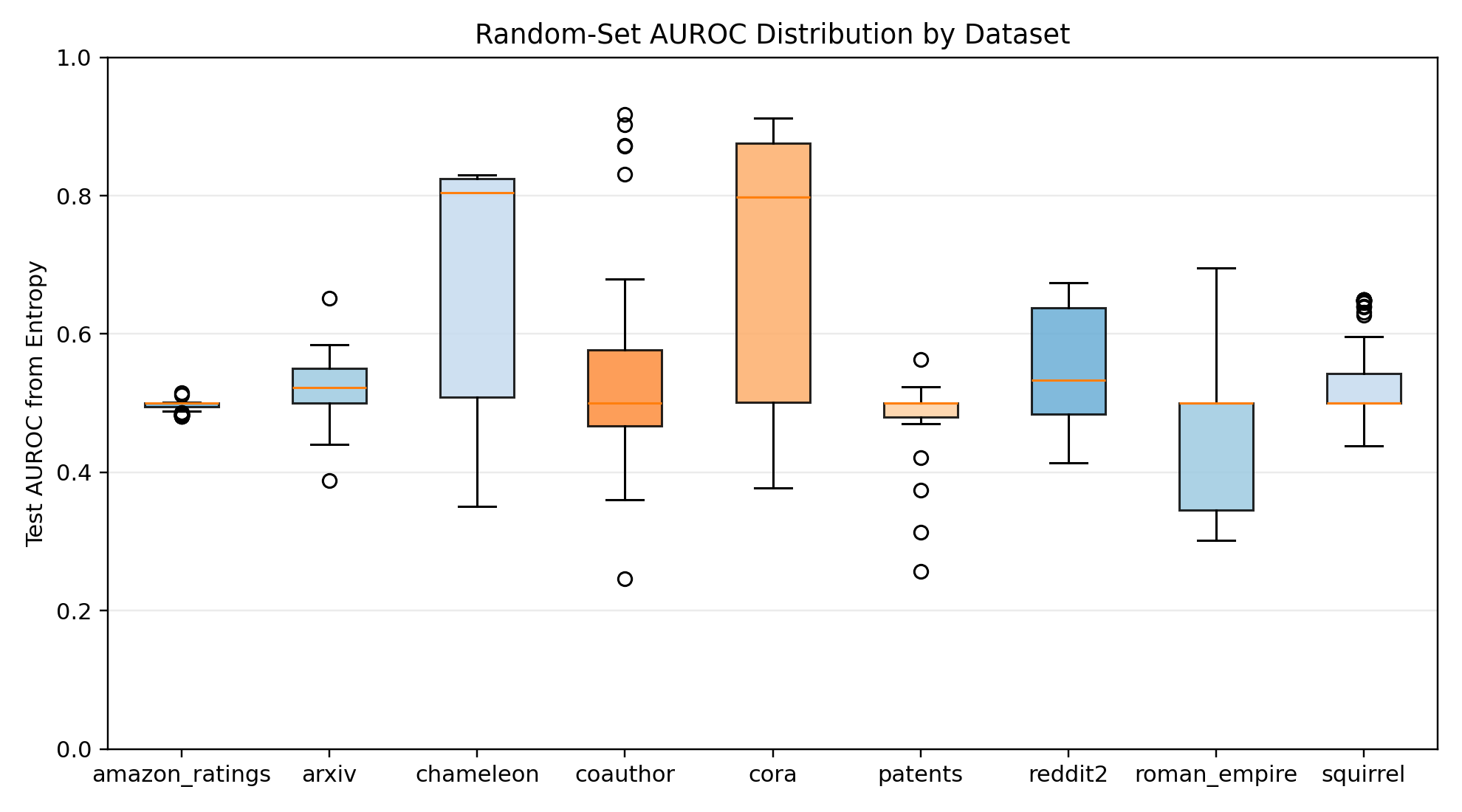}
    \caption{Distribution of RS-GNN AUROC scores across datasets, showing variability in uncertainty-aware performance over multiple runs. The figure highlights both the central tendency and spread of results, illustrating how RS-GNN behaves across diverse graph structures and task settings.}
    \label{fig:dataset_auroc_distribution}
    \end{figure}

The trade-off between ID classification performance and OOD detection quality also emerges clearly from the experiments as represented in \ref{fig:dataset_auroc_distribution}. Across several datasets, including \texttt{patents}, \texttt{reddit2}, and \texttt{roman empire}, the configurations yielding the highest OOD AUROC were not always those yielding the highest ID accuracy. This finding is significant because it shows that optimisation for predictive performance alone is insufficient in uncertainty-aware graph learning. A configuration that appears preferable under conventional classification criteria may not be the one that provides the most reliable uncertainty estimates. Accordingly, the results support the use of joint evaluation criteria when selecting random-set configurations. This is especially relevant in practical settings where uncertainty quality is not secondary, but central to the intended application. \cite{bazhenov_towards_2022}

The dataset-level results also suggest that the usefulness of the random-set formulation depends on the characteristics of the task. On \texttt{patents} and  \texttt{reddit2}, where the scale and complexity of the data make uncertainty estimation particularly relevant, the method provides a meaningful test of whether structured belief representations can improve OOD awareness. On \texttt{cora}, the smaller and more standard benchmark setting serves as a useful baseline, but it is less informative about the behavior of the method under more demanding conditions. Datasets such as \texttt{roman empire}, \texttt{amazon ratings}, and coauthor help bridge this gap by introducing additional diversity in graph topology, label structure, and task difficulty. Similarly, squirrel and chameleon are valuable because they are known to behave differently from more homophilous citation-style graphs, making them useful for testing whether the uncertainty framework remains informative under less conventional graph regimes. \texttt{Arxiv} further contributes by providing a large-scale citation setting distinct from \texttt{cora}, allowing comparison between smaller and larger citation-style benchmarks.

The evidence presented in \ref{fig:best_per_dataset} and \ref{fig:dataset_auroc_distribution} suggests that the random-set model is a credible approach to uncertainty-aware graph learning, particularly when the evaluation explicitly values both predictive performance and uncertainty quality. However, the results also indicate that its effectiveness depends materially on dataset properties, configuration choices, and focal-set construction strategy. For this reason, the method should not be interpreted as uniformly superior across all graph domains. Rather, its contribution lies in offering a structured representation of uncertainty whose strengths become apparent when evaluated across diverse datasets such as \texttt{patents}, \texttt{reddit2}, \texttt{cora}, \texttt{roman empire}, \texttt{squirrel}, \texttt{chameleon}, \texttt{arxiv}, \texttt{coauthor}, and \texttt{amazon ratings}. It's fair to conclude that the value of random sets lies not only in peak performance on a single benchmark, but in the broader insight they provide into how uncertainty can be represented, controlled, and evaluated in graph neural network settings.

\section{Ablation studies}

To further examine the behaviour of RS-GNN, an ablation study was conducted to assess the impact of focal set construction and loss formulation on both predictive performance and uncertainty estimation. Four configurations were considered: singletons-only with full training, standard focal sets with full training, standard focal sets with binary cross-entropy loss, and standard focal sets with BCE and margin regularisation. These configurations allow for a controlled comparison between simpler and more expressive focal set representations, as well as between different optimisation strategies. Ablation experiments were conducted on cora and roman empire rather than the full benchmark suite. This subset was selected to span the principal operating regimes of the random-set model: cora represents a smaller benchmark where full focal-set enumeration is feasible, while roman empire introduces a higher class-count setting in which budgeted focal-set construction is required. This selection allows the contribution of the main random-set design choices to be studied across representative conditions, while keeping the computational cost of the ablation study tractable.

The singletons-only configuration achieves the highest mean AUROC alongside relatively strong average accuracy. This suggests that even a restricted focal set can provide useful uncertainty estimates. However, the corresponding distribution exhibits a higher degree of variability across runs, indicating that this configuration is less stable. In this setting, the model is limited to assigning belief to individual classes, which restricts its ability to explicitly represent ambiguity between classes.

\begin{figure}[!h]
    \centering
    \hspace{-2.3em}
\includegraphics[width=1.05\textwidth]{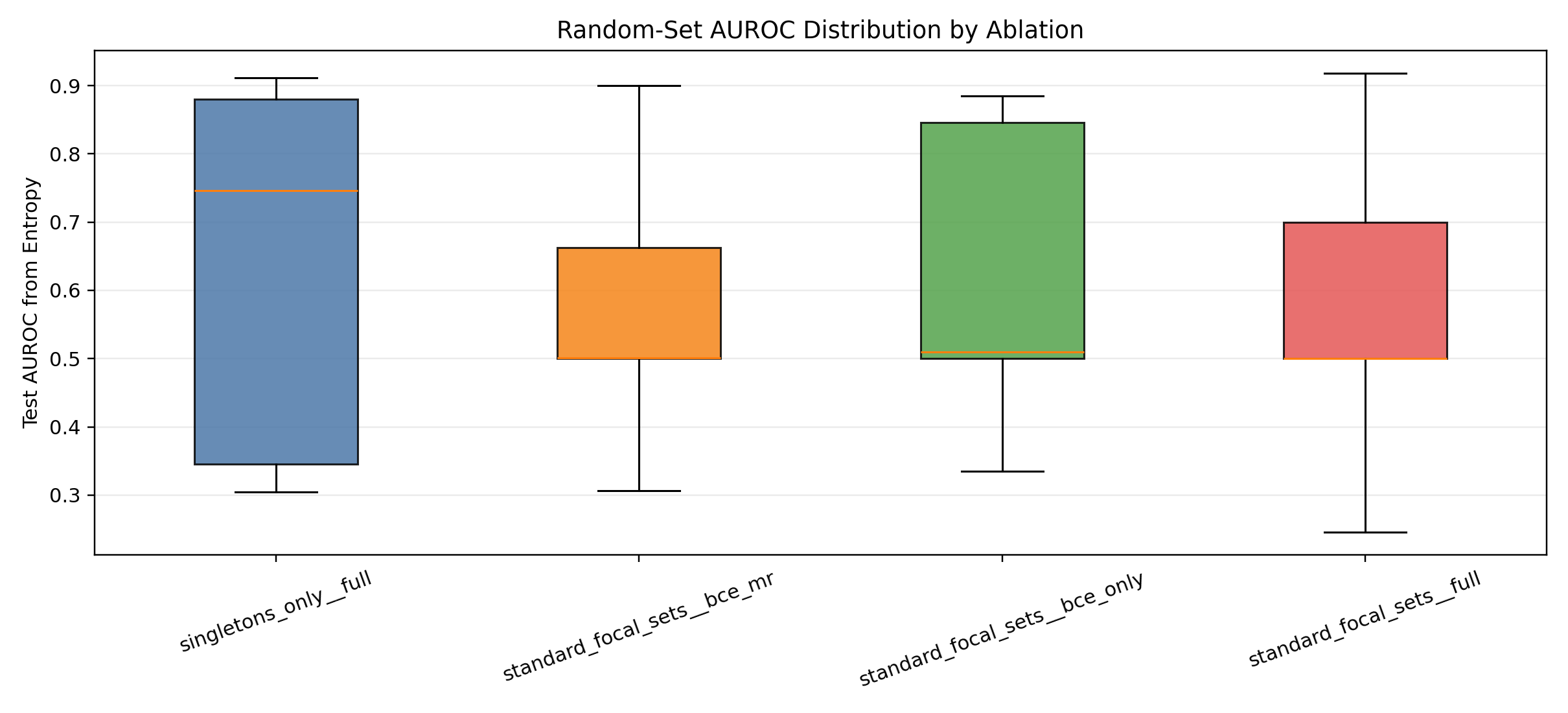}
    \vspace{-15pt}
    \caption{The results indicate that the choice of focal set has a measurable effect on both performance and stability. }
    \label{fig:ablation_auroc_distribution}
    \end{figure}

By contrast, the standard focal set configuration as shown in \ref{fig:ablation_auroc_distribution} achieves the highest observed AUROC, while maintaining a more consistent distribution of results. Although its mean AUROC is slightly lower than that of the singletons-only configuration, the improved stability suggests that the additional expressiveness of the focal set enables more reliable modelling of uncertainty. In particular, the inclusion of multi-class subsets allows the model to capture partial knowledge in cases where evidence does not strongly support a single class.

The BCE-based variants (standard focal sets with BCE and BCE with margin regularisation) demonstrate lower performance overall, with mean AUROC values of 0.6070 and 0.5777, respectively. These results suggest that the choice of loss function has a significant influence on the quality of the learned belief assignments. In particular, optimisation using standard probabilistic loss functions appears less effective in this context, as such losses are not explicitly designed to operate over set-valued representations. In contrast, the full training objective, which directly operates in belief space, is better aligned with the underlying uncertainty framework.

These findings indicate that the effectiveness of RS-GNN depends on learned graph representations, and on the structure upon which uncertainty is expressed. Simpler focal set constructions may yield strong average performance, but more expressive configurations provide improved stability and stronger peak performance. This supports the broader observation that, within RS-GNN, uncertainty estimation is closely tied to the design of the focal set, and should therefore be treated as an integral component of the model rather than a secondary consideration.

\begin{figure}[!h]
    \centering
        \vspace{-15pt}
    \hspace{-2.3em}
\includegraphics[width=1.05\textwidth]{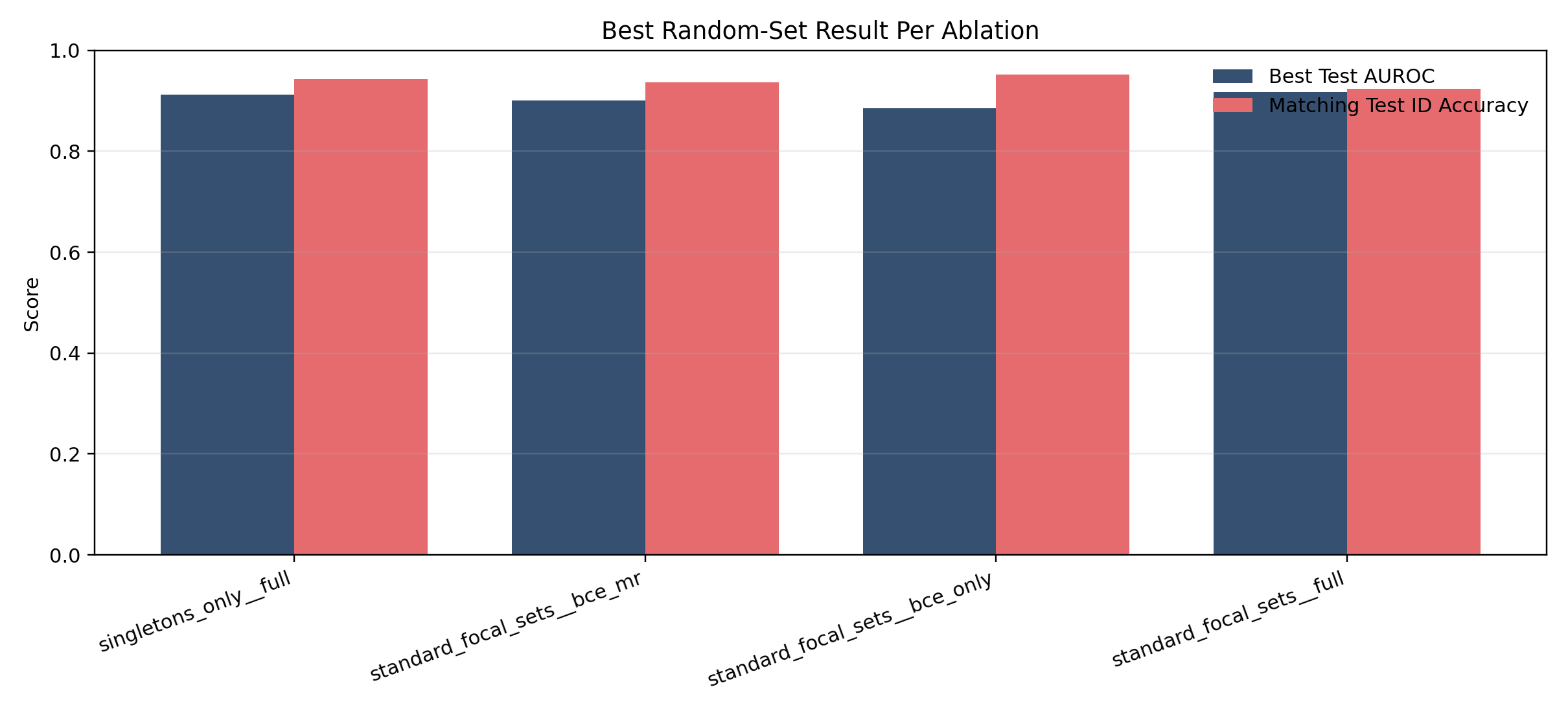}
    \vspace{-15pt}
    \caption{These results allow us to distinguish between AUROC and ID accuracy for the ablation studies performed. }
    \label{fig:best_per_ablation.png}
    \end{figure}

Furthermore, the results in \ref{fig:best_per_ablation.png} highlight the role of focal-set construction in balancing expressiveness and computational efficiency. While configurations using the full focal-set space achieve the highest peak AUROC, the performance gap relative to reduced or singleton-based focal sets remains relatively small. This suggests that the primary gains of RS-GNN can be positively attributed to the combinatorial richness of the focal-set family and from the underlying belief-based representation itself. In practical terms, this indicates that lightweight focal-set configurations may provide a more favourable trade-off between computational cost and uncertainty quality, particularly for large-scale datasets where full enumeration is infeasible. In addition, the comparison between different training objectives reveals that variations such as binary cross-entropy alone and its margin-regularised counterpart produce only marginal differences in performance. This observation reinforces the idea that the effectiveness of RS-GNN is driven more by its representational framework than by specific optimisation choices. While regularisation contributes to stabilising training and ensuring valid mass assignments, it does not fundamentally alter the model’s ability to capture uncertainty.

Another notable pattern is the divergence between classification accuracy and AUROC across configurations. Although test accuracy remains consistently high, AUROC exhibits greater variability, indicating that improvements in predictive performance do not necessarily translate to better uncertainty estimation. This reinforces the need to evaluate uncertainty-aware models using metrics that explicitly capture their ability to distinguish between in-distribution and out-of-distribution samples.

Finally, the distribution of results across runs suggests differences in stability between configurations. While several setups achieve comparable peak performance, their mean AUROC values differ more noticeably, indicating that some configurations are more sensitive to initialisation and training dynamics. This highlights the importance of considering not only best-case results but also consistency when evaluating uncertainty-aware models.

In conclusion, this ablation study has demonstrated that the effectiveness of RS-GNN is primarily attributable to its belief-function representation, with focal-set design and training objective playing secondary but still meaningful roles. These findings support the broader idea that structured uncertainty representations can provide robust performance across different configurations, while also allowing flexibility in adapting the model to varying computational and dataset constraints.

\section{Conclusions}

In this paper, we proposed an original new framework in which node-level epistemic uncertainty is modelled in a belief function formalism. 
The resulting Random-Set Graph Neural Networks employ a belief-function head to predict a finite random set over the list of classes, from which both a precise probability prediction (via the pignistic transform) and a measure of epistemic uncertainty (expressed by the size of the predicted probability intervals for each class) can be obtained. 
Extensive experiments on 9 different graph learning datasets, including challenging real-world autonomous driving benchmarks as such Nuscene and ROAD, demonstrate RS-GNN's superior uncertainty quantification capabilities.

The natural next step is the full integration of belief function representations in the message passing core of the GNN, building on existing body of knowledge on belief function propagation. 

\bibliographystyle{unsrtnat}
\bibliography{bib}

\end{document}